# iProStruct2D: Identifying protein structural classes by deep learning via 2D representations

**Loris Nanni[1], Alessandra Lumini[2], Federica Pasquali[1], Sheryl Brahnam[3*]**

[1] Information Engineering, University of Padua, Via Gradenigo 6, 35131 Padova, Italy
{loris.nanni, Federica.pasquali}@unipd.it
[2] Computer Science and Engineering, Università di Bologna, Via dell'Università 50, 47521 Cesena, Italy.
alessandra.lumini@unibo.it
[3] Information Technology and Cybersecurity, Glass Hall 387, Missouri State University, 901 S. National, Springfield, MO 65804, USA.
sbrahnam@missouristate.edu

*Abstract*—In this paper we address the problem of protein classification starting from a multi-view 2D representation of proteins. From each 3D protein structure, a large set of 2D projections is generated using the protein visualization software Jmol. This set of multi-view 2D representations includes 13 different types of protein visualizations that emphasize specific properties of protein structure (e.g., a backbone visualization that displays the backbone structure of the protein as a trace of the $C_\alpha$ atom). Each type of representation is used to train a different Convolutional Neural Network (CNN), and the fusion of these CNNs is shown to be able to exploit the diversity of different types of representations to improve classification performance. In addition, several multi-view projections are obtained by virtually rotating the protein structure around its central X, Y, and Z viewing axes to produce 125 images. This approach can be considered a data augmentation method for improving the performance of the classifier and can be used in both the training and the testing phases. Experimental evaluation of the proposed approach on two datasets demonstrates the strength of the proposed method with respect to the other state-of-the-art approaches. The MATLAB code used in this paper is available at https://github.com/LorisNanni.

## I. INTRODUCTION

Protein Structure Comparison (PSC) is an essential task in structural biology and drug discovery; it allows researchers, for instance, to infer protein evolution (to understand better the relationship between protein structure and function) and to transfer knowledge about known proteins to a novel protein (Schenkel, Holm, Rosenström, & Kääriäinen, 2008). Some of the main applications of PSC include establishing structural, evolutionary, and functional relationships between proteins; assigning functional annotations to proteins (Mills, Beuning, & Ondrechen, 2015); drug repositioning (Haupt, Daminelli, & Schroeder, 2013); and identification of proteins with similar binding sites as potential targets for the same ligand (Duran-Frigola et al., 2017).

In general PSC methods try to provide a measure of structural similarity between proteins that can be used to identify evolutionarily related proteins and similar folds. The first step in many such measures includes an alignment process that is based on the definition of residues having structurally equivalent roles in the proteins being compared (structure alignment is considered more reliable than sequence alignment since during evolution the conservation of protein structure is stronger).

The second step after alignment is the search for geometrical transformations that minimize the distance between residues (Bourne & Shindyalov, 2005). The evolution of proteins includes insertions, deletions, and mutations of single residues, exon shuffling, and gene fusion and duplication (Russell et al., 1997). In general, such changes mainly affect the surface regions of the proteins, while the functional sites tend to be maintained if the protein retains the same molecular function.

The last step defines a distance measure that ideally quantifies the structural differences among proteins. An ideal measure should produce a single number in a fixed range [e.g. 0-1], be able to distinguish between related and nonrelated structural pairs, be able to capture the nature of protein folding or protein interaction determinants, and be robust against minor changes in structures (Kufareva & Abagyan, 2012).

In the literature, several structural alignment methods and structural distance measures have been proposed. Classical methods in protein structural class prediction include (Chou, 1995; C. T. Zhang & Chou, 1994; G. P. Zhou, 1998). Some popular approaches for protein alignment that are available to the structural biologist on the protein databank (PDB) website include (Novotny, Madsen, & Kleywegt, 2004): DALI (Distance matrix ALIgnment) (Holm & Sander, 1993), CE (Combinatorial Extension) (Shindyalov & Bourne, 1998) and FATCAT (Ye & Godzik, 2003). These methods compare the geometry of the $C_\alpha$ backbone atoms based on different algorithms and work best when the sequence identity of the protein is high, but they often have difficulties when protein structures are very dissimilar. This difficulty is based on determining a single optimal alignment of functional similarity and/or tertiary structure similarity and is magnified whenever the sequence identity of the protein is less than 20%, the point at which structural differences become very large (Chothia & Lesk, 1986). Another strategy for protein alignment takes into account not only the backbone geometry but also the physicochemical environment of each residue: for example MATRAS (Kawabata, 2003) performs a first alignment by matching secondary structure elements, and then by considering environmental properties and $C_\alpha$ distances to refine the solution.

As far as the distance measure is concerned, the most commonly used quantitative measure of similarity between protein structures is the Root Mean Square Deviation (RMSD) (Kufareva & Abagyan, 2012), which is calculated between the superimposed $C_\alpha$ chains. RMSD suffers from amplitudes of errors, however, mainly in cases when two structures cannot be effectively superimposed or when the length of the alignment is large (Ye & Godzik, 2003). Several different measures have been proposed to overcome this lack of refinement. The weighted RMSD (wRMSD) (Kufareva & Abagyan, 2012), for example, gives different weights to selected atomic subsets which softens the unstructured regions. The Global Distance Test (GDT) and Longest Continuous Segment (LCS) measures (Kufareva & Abagyan, 2012) are based on the selection of the largest set of the model residues that can be superimposed with the corresponding set in the reference structure, and the TM-score (Y. Zhang & Skolnick, 2005) applies a weight-based measure on the length of the aligned structure, thus avoiding dependence of the obtained score on the target size.

Another class of approaches is based on a global 3D representation of the protein: the main idea here is to base similarity between structures (and not on alignment, as with local approaches) by comparing descriptors extracted from the 3D geometrical structures (or their 2D projections) (Harder, Borg, Boomsma, Røgen, & Hamelryck, 2012; Mirceva, Cingovska, Dimov, & Davcev, 2012; Sael et al., 2008; X. Zhou, Chou, & Wong, 2006). Some noteworthy global approaches include the work in (Mirceva et al., 2012) where wavelet-based protein descriptors are proposed, the work in (Sael et al., 2008) where the 3D geometrical surface of the protein structures are represented by 3D Zernike descriptors, and the approaches proposed in (Harder et al., 2012) where Gauss integral vectors are extracted from protein structures represented as open curves.

The third class of methods recently proposed by (Suryanto, Saigo, & Fukui, 2015) aims at overcoming the difficulty related to the alignment process and the instability of measurements inherent in both local-alignment and global-representation methods by representing a protein as a set of 2D multi-views of its 3D molecular structure. By using a protein visualization software package, it is possible to inspect the protein structures from multiple viewpoints and according to different types of representations. The basic idea behind the 2D view-based approaches is to use projections onto the 2D plane of the computer screen obtained by analyzing a 3D structure from different points of view obtained by performing several 3D rotations of the structure. The approach in (Suryanto et al., 2015) uses texture descriptors to define different protein subspaces via a measure of similarity known as the Mutual Subspace Method (MSM) (Maeda, 2010) for evaluating the similarity between protein structures. It is then possible to combine this type of representation with several state-of-the-art image descriptors and general-purpose classifiers (Nanni, Lumini, & Brahnam, 2014).

Even though 2D representations do not provide any real knowledge regarding the "binding pockets" of proteins with their ligands, knowledge which is useful, for instance, for designing therapeutic drugs (Chou, Tomaselli, & Heinrikson, 2000), we recognize that using 2D views of a protein, similar to those proposed in (Suryanto et al., 2015), can be yet another valuable way for solving the protein classification problem using Deep Learning (DL). DL is exceptionally effective when applied to image classification tasks. The most studied DL architecture for image classification is the Convolutional Neural Network (CNN) (Gua et al., 2018), a multi-layered neural network inspired by the natural visual perception mechanism of the human being.

Some researchers have also designed CNN models to classify 3D shapes directly from 3D representations (Wu et al., 2015), but other studies, such as the ensemble proposed in (Su, Maji, Kalogerakis, & Learned-Miller, 2015), have shown that building classifiers of 3D shapes from 2D image renderings makes for a more feasible solution. A multi-view CNN architecture is trained to recognize objects from their rendered 2D views. The ensemble in (Su et al., 2015), for instance, which differs from the one proposed in this work, has a different network for each 2D view.

In this work, we try to show that it is possible to use DL to perform protein classification starting from 2D snapshots taken from 3D structures of proteins rendered using protein simulation/drawing software, such as Jmol (Murzin, Brenner, Hubbard, & Chothia, 1995). Our system for protein classification takes pretrained CNN models and fine-tunes them on a set of multi-view 2D images of 3D protein structures. A set of a selection of 2D-views is treated as a form of data augmentation for the input data. Unlike (Suryanto et al., 2015) which considers only four protein representations, in this work we evaluate 13 protein representations. We then select those that performed best to train different CNNs that are finally fused together.

The main objective in this work is to show that treating the different views of a protein as a type of data augmentation improves the performance of a CNN classifier, and the choice of different representations increases classifier diversity. Ensembles of networks trained using these different types of representations are tested on two benchmark datasets (Suryanto et al., 2015), and their performance is compared with state-of-the-art methods published in the literature. Results demonstrate the strength of the method proposed in this paper.

## II. PROPOSED METHOD

As noted in the introduction, we utilize many types of visualization to fine-tune pretrained CNNs for the problem of classifying protein structures. In this section, we outline our proposed method. We begin in section A by reviewing the basics of CNN and the pretrained CNN models used in this paper. Then, in section B, we discuss the use of protein visualization software to generate different types of visualizations/images for representing a protein. Finally, in section C, we propose a set of 2D multi-views of a given protein as a form of data augmentation for improving the fine-tuning of CNN.

A schematic of the proposed approach is provided in Fig. 1 where the input protein in PDB format is rendered in 3D (using a 3D molecular graphics program) into 13 different representations. For each representation, a different CNN is fine-tuned taking 125 2D views as the input (other values are examined in the experimental section). In the testing phase, the classification results obtained by the different views and representations are averaged to produce the classification score.

### A. Fine-tuning convolutional neural networks

CNN (Gua et al., 2018) is a multi-layered neural network that incorporates spatial context and weight sharing between pixels to learn the optimal image descriptors and weights for a given image classification task.

A CNN is a feed-forward network built with repeated combinations of different types of layers: *convolutional layers*, aimed at convolving input to filters, *activation layers*, used to introduce nonlinearity to the system, *pool layers*, which perform downsampling, *fully-connected layers*, which simulate neural connections, and *classification layers*, which perform the final classification. CNNs are trained using backpropagation: given a first, possibly random, initialization, a forward propagation step is performed to find the classification errors, then backpropagation is used to calculate the gradients of the error with respect to all the weights in the network. Due to the large number of parameters, the training phase of CNN is quite time-consuming and requires a large dataset to avoid overfitting. In applications where the training set is not large enough to train from scratch, transfer learning (Yosinski, Clune, Bengio, & Lipson, 2014) has proven useful.

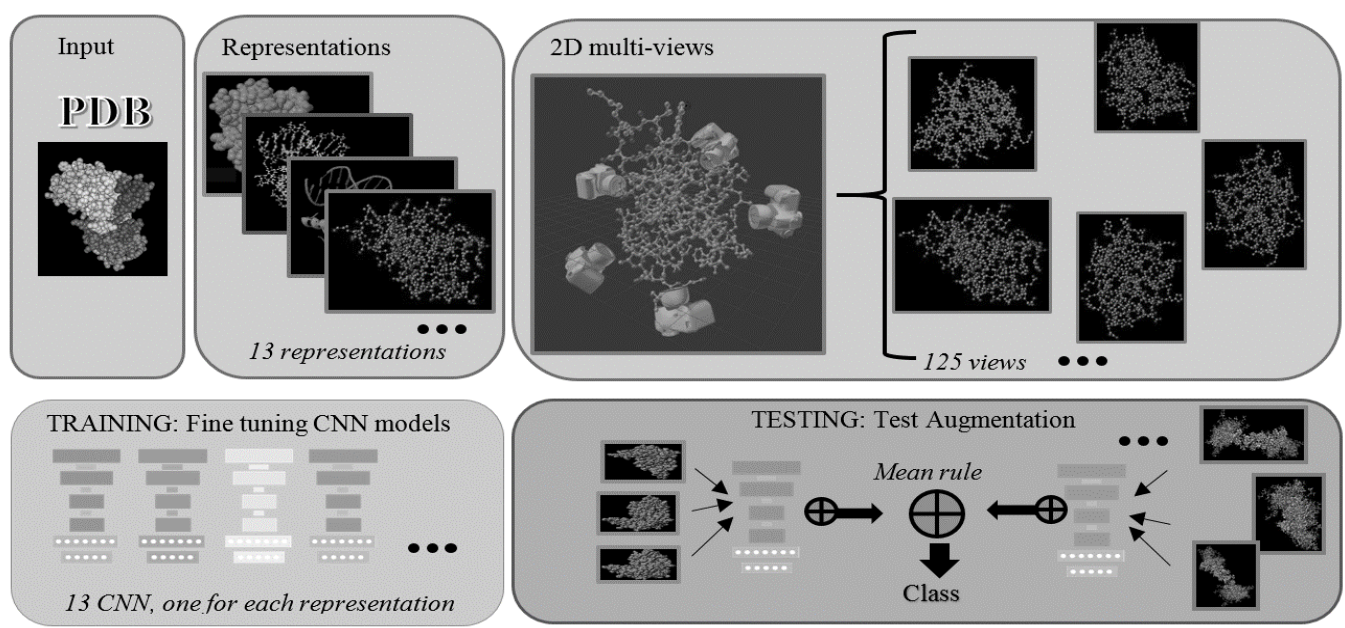

Fig. 1. Schematic of the proposed approach: given an input PDB, 125 2D views are extracted from 13 types of representation. In the training phase, the images are used to fine-tune two pretrained CNNs. In the testing phase, the scores from the CNNs are averaged to perform classification.

Several different CNN models have been proposed that have been pretrained on enormous image datasets and shared with the research community: AlexNet (Krizhevsky, Sutskever, & Hinton, 2012), VGGNet (Simonyan & Zisserman, 2014), GoogleNet (Szegedy et al., 2015) and ResNet (He, Zhang, Ren, & Sun, 2016) being some of the most famous. In this work, we perform fine-tuning on three pretrained models: AlexNet, GoogleNet, and ResNet50.

To perform fine-tuning, we alter the last fully connected and classification layers so that they match the number of classes, without freezing the weights of the previous layers. The performance reported in section III is obtained using the following tuning parameters: number of epochs is set to 20, mini-batch size to 30, and we use a fixed learning rate of 0.001.

### B. Protein visualization

Nowadays, there are many 3D molecular graphics programs (O'Donoghue et al., 2010) that allow users to visualize the structure of proteins. In this paper, we focus on the 3D visualization of a protein's PDB code. This approach highlights in an accurate and immediate way a protein's structure and principal characteristics. For this study, Jmol (Murzin et al., 1995) was used for producing the 3D representations of the protein's PDB code. Jmol is a free, open-source Java viewer for molecular visualization that provides different classes of visualization: *Display Format*, used for choosing how to visualize atoms and bonds; *Display Structure*, used to show the structure of the protein in different ways; *Display Color*, used for highlighting specific characteristics of the molecule, and *Display Thickness*, used to fix the thickness of the structure.

A summary of the different visualizations offered by each class (see Fig. 2) are as follows:

*1) Display Format*:

- BALL&STICK: the default display format option, where atoms are represented by spheres connected by bars that represent the bonds and where the color of every sphere points out the corresponding chemical element (the colors of the spheres follow the CPK scheme).
- SPACEFILL: where every atom is represented by a sphere whose radius is proportional to the radius of the atom; Atoms of different chemical elements are represented by spheres of different colors.
- WIREFRAME: where bonds among the atoms are represented as cylindrical sticks with a fixed diameter, while the extremities of the cylindrical sticks point out the position of the atoms.

*2) Display Structure*:

- BACKBONE: an option that illustrates the secondary structures inside a molecule with a zigzag line that is drawn so that it connects the main atoms in the backbone alpha carbons in a protein and phosphorus atoms in a nucleic acid.
- CARTOON: where molecules are represented as ribbons in stretches in which the alpha helixes or beta sheets are present; each stretch ends with an arrowhead.
- RIBBONS: an option like BACKBONE except that it displays the line that connects the main atoms in the backbone (alfa carbons in a protein and phosphorus atoms in a nucleic acid) as a solid flat ribbon.
- ROCKETS: an option that assigns cylinders in stretches in which the alpha helixes and planks for beta stretches are present; both end with an arrowhead.
- STRANDS: an option like RIBBONS and CARTOONS that displays the backbones as a series of thin lines so that the molecular structure is represented by parallel longitudinal threads.
- TRACE: an option that is analogous to BACKBONE, except that it draws a smooth curve passing through the middle points between successive atoms in the alpha carbons of a peptide chain or the phosphorus atoms of nucleic acids.

*3) Display color*:

- AMINO: an option that assigns to each of the 20 standard amino acids a fixed color depending on its chemical properties: bright colors for those that are polar and dark colors for those that are hydrophobic.
- CHAIN: an option that assigns a different color to every chain of the structure in the case where a PDB file is composed of more than one chain. This option is beneficial in the analysis of protein structure.
- CHARGE: an option that assigns a color to every atom depending on its charge, using a gradient of color from red (negative charge) to white (zero charge) to blue (positive charge).
- CPK: the default color option, where a color is assigned to each element according to the following scheme: Carbon is gray, Hydrogen is white, Oxygen is red, Nitrogen is blue. and Sulfur is yellow.
- STRUCTURE: an option that underlines each secondary structure of a protein in a different way: the alpha-helixes are colored fuchsia, the beta sheets yellow, and all the others are white. DNA and RNA are shown in purple and red. This format is very useful when combined with a *Display Format* visualization that makes evident the secondary structure of the molecule in the examination as BACKBONE or CARTOON.

*4) Display Thickness*:

- Jmol can vary the thickness of the Display Format and Display Structure (measured in Angstrom units or RasMol units).

In this work we obtained 13 different types of visualizations (see Fig. 2) by combining the different types of display parameters described above: *SPACEFILL, WIREFRAME,* and *BALL&STICK* are obtained using the respective display formats; *AMINO, CHAIN, CHARGE* and *STRUCTURE* are variations of the *BALL&STICK* format that change colors (from the default, CPK); and *BACKBONE* to *TRACE* are the display structures described above. All images are obtained by fixing the display thickness as SPACEFILL (200), WIREFRAME (60), BACKBONE (150), STANDS (300), and TRACE (300).

*C.  2D multi-view generation*

The 2D multi-view protein images can be obtained by uniformly rotating each of the 3D protein structures around their central X, Y, and Z axes. In our system, we generate uniform rotation angles with steps of 45° from 0° to 180° in each axis; therefore, the total number of 2D projections from each type of visualization is 125.

All the generated poses are used both for data augmentation in the training phase and for "testing time augmentation," which means that to classify an unknown structure all 125 poses are generated and then predictions are calculated for all these images, taking the average score as the final prediction.

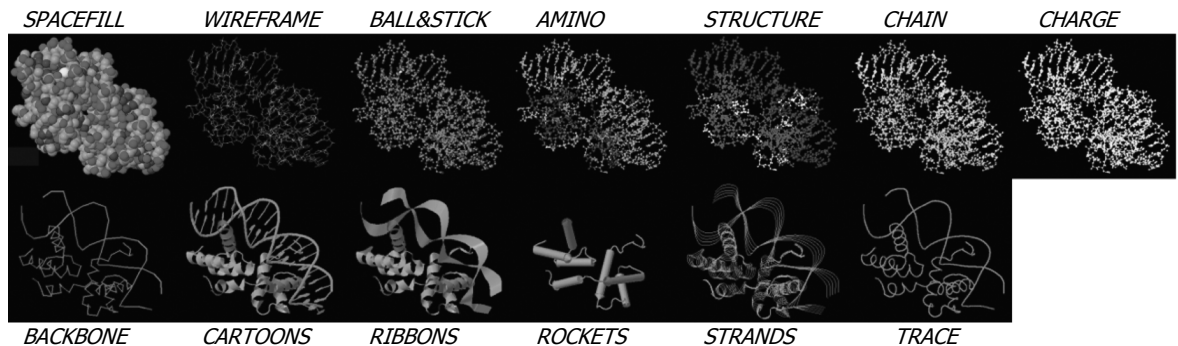

Fig. 2. Visualization of different types of protein representations generated using JMol on the PDB 1akh (color available with e-publication).

## III. EXPERIMENTAL RESULTS

To evaluate the effectiveness of the proposed approach, we evaluate its classification performance on the two datasets used in (Suryanto et al., 2015) (available at http://doi.ieeecomputersociety.org/10.1109/TCBB.2016.2603987).

**Fold95**: this dataset for protein fold classification includes 95 protein structures that have no more than 10% sequence identity. The proteins are divided into five classes according to their fold similarity: (1) f.1 toxins' membrane translocation domains (9 proteins); (2) f.17 transmembrane helix hairpin (9 proteins); (3) f.21 heme-binding four-helical bundle (9 proteins); (4) f.23 single transmembrane helix (50 proteins); and (5) f.4 transmembrane beta-barrels (18 proteins). Due to the imbalance of classes (each class includes from 9 to 50 proteins), stratified random sampling is used (Suryanto et al., 2015) to feed the data into 10-fold cross-validations.

**Class700**: this is a dataset for protein class classification. It contains 700 proteins with at most 20% sequence identity. Images are equally distributed among the seven classes (100 proteins from each class) of the SCOP protein classification scheme (Murzin et al., 1995): (1) α-proteins (containing mainly α-helices); (2) β-proteins (containing mainly β-sheets); (3) α/β-proteins (containing both α and β structures where the β-sheets are parallel); (4) α+β-proteins (containing both α and β structures where the β-sheets are anti-parallel); (5) multi-domain proteins that have multi-functions; (6) membrane and cell surface proteins; and (7) small proteins. A 10-fold cross-validation protocol is used that maintains the distribution among classes.

TABLE I
AUC OF DIFFERENT CNN IN Fold95 AND Class700 DATASETS

|  | Fold95 | | | Class700 | | |
|---|---|---|---|---|---|---|
|  | AlexNet | GoogleNet | ResNet | AlexNet | GoogleNet | ResNet |
| BALL&STICKS | 95.6 | 97.7 | --- | 85.4 | --- | --- |
| AMINO | 95.5 | 98.1 | --- | 85.0 | --- | --- |
| CHAIN | 90.6 | 91.3 | --- | 79.5 | --- | --- |
| CHARGE | 92.3 | 94.7 | --- | 82.4 | --- | --- |
| STRUCTURE | 84.9 | 96.2 | --- | 90.8 | --- | --- |
| SPACEFILL | 64.5 | 96.2 | --- | 82.4 | --- | --- |
| WIREFRAME | 96.7 | 96.5 | --- | 87.1 | --- | --- |
| BACKBONE | 94.8 | 96.3 | --- | 93.5 | --- | --- |
| CARTOON | 98.7 | 97.8 | --- | 93.9 | --- | --- |
| RIBBONS | 98.3 | 97.5 | 97.3 | 94.2 | 94.7 | 95.4 |
| ROCKETS | 94.7 | 96.7 | 96.1 | 93.7 | 94.6 | 95.5 |
| STRANDS | 98.0 | 97.7 | 97.5 | 94.3 | 94.3 | 95.5 |
| TRACE | 85.3 | 97.0 | --- | 92.3 | --- | --- |
| TOP2 | 98.4 | 97.9 | 97.5 | 94.8 | 94.8 | 95.7 |
| TOP3 | 98.5 | **98.6** | 97.9 | 94.8 | **95.2** | **95.8** |
| TOP3b | 98.5 | --- | ---- | 94.8 | --- | --- |
| TOP4 | 98.5 | --- | ---- | 94.9 | --- | ---- |
| ORACLE | 99.2 | 99.5 | 99.0 | 96.8 | 97.5 | 97.7 |

For internal evaluation, we use the area under the ROC curve (AUC) as the performance indicator (Fawcett, 2004). To compare results with the literature in those cases where AUC is not reported, we used Accuracy.

The first experiment is related to the classification of proteins using the CNN models described in section II, fine-tuned using the different representation types. In Table I the classification results are reported, along with the performance of some fusion approaches. Due to computational issues, GoogleNet and ResNet are tuned only for the most performing representations.

In the rows labeled TOP$x$, we report the performance obtained by the sum rule fusion of the following representations, which obtained the best stand-alone performance:

- TOP2: the fusion by sum rule of RIBBONS and STRANDS;
- TOP3: the fusion by sum rule of RIBBONS, ROCKETS, and STRANDS;
- TOP3b: the fusion by sum rule of RIBBONS, CARTOON, and STRANDS;
- TOP4: the fusion by sum rule of RIBBONS, ROCKETS, CARTOON, and STRANDS.

In the last line of Table 1, we report the results obtained by an ORACLE ensemble. The Oracle concept (Kuncheva, 2002) is a hypothetical dynamic selection approach that always selects the classifier that correctly classifies the test sample, if such a classifier exists. The comparison against the ORACLE performance is useful for validating the composition of the ensemble.

In both datasets the highest performance is obtained by TOP3. As can be seen in Table 1, combining different visualizations clearly improves performance, and different visualizations emphasize different characteristics of the protein.

In Table II we compare our best ensembles with some of the most effective approaches proposed in the literature (reporting both AUC and accuracy):

**AC**: the autocovariance (AC) approach (Zeng et al., 2009) is a sequence-based variant of Chou's PseAAC.

**QRC**: Quasi Residue Couple (QRC) (Nanni & Lumini, 2006) is a feature extraction method for the primary sequence of a protein. The QRC descriptor is calculated by selecting a physicochemical property and combining its values with each nonzero entry in the residue couple.

**TXT**: is based on handcrafted descriptors extracted from PSSM (Position Specific Scoring Matrix) and DM matrix representations of the protein, see (Nanni et al., 2014).

**FATCAT** (Ye & Godzik, 2003) is a method for structural alignment of proteins based on optimizing the alignment and minimizing the number of rigid-body movements (twists) around pivot points (hinges) introduced in the reference protein.

**CE** (Shindyalov & Bourne, 1998) is an alignment algorithm which involves a combinatorial extension (CE) of an alignment path defined by aligned fragment pairs (AFP).

**TM** (Y. Zhang & Skolnick, 2005): TM-align is an algorithm for protein structure comparisons based on optimized residue-to-residue alignment.

**GDA** (Suryanto et al., 2015) is an approach based on multi-view 2D images of 3D protein structures.

**TOP3-G+R** is the combination by sum rule of our two best ensembles (TOP3-GoogleNet+TOP3-ResNet).

TABLE II
COMPARISON WITH THE LITERATURE IN FOLD95 AND CLASS700 DATASETS: BOTH AUC AND ACCURACY ARE USED AS PERFORMANCE INDICATORS

| | Fold95 | | Class700 | |
|---|---|---|---|---|
| | Accuracy | AUC | Accuracy | AUC |
| AC (Zeng et al., 2009) | 0.711 | 84.00 | 0.471 | 80.83 |
| QRC (Nanni & Lumini, 2006) | 0.677 | 82.37 | 0.432 | 78.05 |
| TXT [24] | 0.822 | 93.60 | 0.628 | 88.39 |
| FATCAT (Ye & Godzik, 2003) | - | - | 0.531 | - |
| CE (Shindyalov & Bourne, 1998) | - | - | 0.491 | - |
| TM (Y. Zhang & Skolnick, 2005) | 0.863 | - | 0.640 | - |
| GDA (Suryanto et al., 2015) | 0.884 | - | 0.694 | - |
| TOP3 - AlexNet | **0.922** | 98.5 | 0.739 | 94.8 |
| TOP3 - GoogleNet | 0.878 | **98.6** | 0.754 | 95.2 |
| TOP3 - ResNet | 0.911 | 97.9 | 0.769 | **95.8** |
| TOP3 - G+R | 0.900 | 98.5 | **0.771** | 95.6 |
| ORACLE | 0.967 | | **0.870** | |

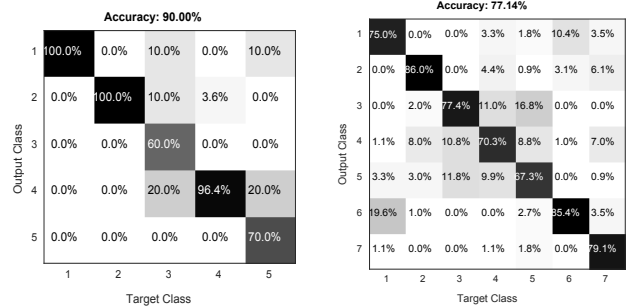

Fig. 3. Confusion matrices of TOP3-G+R in Fold95 (left) and Class700.

TABLE III
AUC OBTAINED VARYING THE NUMBER OF IMAGES USED FOR DESCRIBING EACH PROTEIN USING ALEXNET ON STRANDS.

| Number of Images: | 30 | 60 | 125 | 200 |
|---|---|---|---|---|
| Fold95 | 94.9 | 97.5 | 98.0 | 98.1 |
| Class700 | 90.2 | 93.0 | 94.3 | 94.3 |

From the results in Table II, we can draw some interesting conclusions. The proposed representation based on multi-view 2D images is well suited to protein classification, as proved by the good performance of the proposed approach and GDA (Suryanto et al., 2015) with respect to older representations like the amino-acid sequence (AC, QRC) or PSSM matrix (TXT) or to measures based on protein alignment (FATCAT, CE, TM). The use of different types of representation in the generation of multi-view 2D images makes our approach based on CNN stronger: the ensemble of different representational types outperforms all the other approaches in the literature and works well in both the classification problems.

In Fig. 3 the confusion matrices obtained by our best ensemble in both datasets are reported. In Class700 there is not a strong difference in accuracy among the different classes. In Fold95 the main problems are in class (3) "f.21 heme-binding four-helical bundle" and class (5) "f.4 transmembrane beta-barrels".

As a further experiment intended to study the dependence of performance on the number of 2D views, we report in Table 3 the performance of AlexNet on the STRANDS representation obtained by varying the number of images built for describing each protein. Clearly, using 30 or 60 images decreases performance; whereas using 200 image per protein does not improve performance with respect to using 125 images.

Finally, Table 4 reports training and test time obtained using a Titan Xp NVIDIA for different CNN models. The training time is related to a single fold and representation, and the test time is given for all 125 images that represent a given protein. Clearly, a protein can be classified by the ensemble in a few seconds.

TABLE IV
TESTING AND TRAINING TIMES

| | | AlexNet | GoogleNet | ResNet |
|---|---|---|---|---|
| Training time | Fold95 | 26 min 14 sec | 82 min 33 sec | 121 min 36 sec |
| | Class700 | 199 min 48 sec | 609 min 32 sec | 1472 min 21 sec |
| Test time | | 0.14 sec | 0.16 sec | 0.55 sec |

## IV. CONCLUSION

In this paper our aim has been to show that multi-view 2D representations of proteins can be used to fine-tune a pretrained CNN for a protein classification problem. Preliminary experiments on two datasets have shown the effectiveness of the proposed approach for protein classification: the proposed ensemble obtains a performance that is higher than other state-of-the-art methods.

We are aware that a broader evaluation is needed; therefore, we plan to test our approach with other datasets in the future. Though we have tested 13 different types of representation, we have yet to investigate how to select the most discriminative for a given classification problem. We consider this as a future work as well. Automatic selection may even be more critical if we consider additional types of representation by changing other parameters in the protein visualization software (e.g. thickness).

We also plan on investigating pre-trained CNNs as feature extractors for generating a compact descriptor for a protein: this choice could allow the use of the simple Euclidean distance as the similarity measure between two proteins. Finally, we want to test attention-aware deep networks (Dong & Shen, 2018; Dong et al., 2018; Wang & Shen, 2018) for cropping the most essential part of the protein to reduce noise, which can reduce classification performance.

**Acknowledgments.** We would like to acknowledge the support that NVIDIA provided us through the GPU Grant Program. We used a donated TitanX GPU to train the CNNs used in this work.

REFERENCES

Bourne, P. E., & Shindyalov, I. N. (2005). Structure Comparison and Alignment. In P. E. Bourne & H. Weissig (Eds.), *Structural Bioinformatics* (pp. 321-337).

Chothia, C., & Lesk, A. M. (1986). The relation between the divergence of sequence and structure in proteins. *The EMBO Journal, 5*(4), 823-826. doi:10.1002/j.1460-2075.1986.tb04288.x

Chou, K.-C. (1995). A novel approach to predicting protein structural classes in a (20-1)-D amino acid composition space. *Proteins, 21*, 319-344.

Chou, K.-C., Tomaselli, A. G., & Heinrikson, R. L. (2000). Prediction of the tertiary structure of a caspase-9/inhibitor complex. *FEBS letters, 470*(3), 249-256. doi:10.1016/s0014-5793(00)01333-8

Dong, X., & Shen, J. (2018). *Triplet Loss in Siamese Network for Object Tracking*. Paper presented at the Computer Vision – ECCV 2018.

Dong, X., Shen, J., Wang, W., Yu, L., Shao, L., & Porikli, F. M. (2018). Hyperparameter Optimization for Tracking with Continuous Deep Q-Learning. *2018 IEEE/CVF Conference on Computer Vision and Pattern Recognition*, 518-527.

Duran-Frigola, M., Siragusa, L., Ruppin, E., Barril, X., Cruciani, G., & Aloy, P. (2017). Detecting similar binding pockets to enable systems polypharmacology. *PLOS Computational Biology, 13*(6), e1005522. doi:10.1371/journal.pcbi.1005522

Fawcett, T. (2004). *ROC graphs: Notes and practical considerations for researchers*. Retrieved from Palo Alto, USA:

Gua, J., Wang, Z., Kuen, J., Ma, L., Shahroudy, A., Shuai, B., . . . Tsuhan, C. (2018). Recent advances in convolutional neural networks. *Pattern Recognition, 77*, 354-377. doi:https://doi.org/10.1016/j.patcog.2017.10.013

Harder, T., Borg, M., Boomsma, W., Røgen, P., & Hamelryck, T. (2012). Fast large-scale clustering of protein structures using Gauss Integrals,. *BioInformatics*, 510-515. doi:doi:10.1093/bioinformatics/btr692

Haupt, V. J., Daminelli, S., & Schroeder, M. (2013). Drug Promiscuity in PDB: Protein Binding Site Similarity Is Key. *PLoS ONE, 8*(6), e65894. doi:10.1371/journal.pone.0065894

He, K., Zhang, X., Ren, S., & Sun, J. (2016). *Deep residual learning for image recognition*. Paper presented at the 2016 IEEE Conference on Computer Vision and Pattern Recognition (CVPR), Las Vegas, NV.

Holm, L., & Sander, C. (1993). Protein structure comparison by alignment of distance matrices. *Journal of Molecular Biology, 233*, 123-138. doi:doi:10.1006/jmbi.1993.1489

Kawabata, T. (2003). MATRAS: a program for protein 3D structure comparison. *Nucleic Acids Research, 31*(13), 3367-3369. doi:10.1093/nar/gkg581

Krizhevsky, A., Sutskever, I., & Hinton, G. E. (2012). ImageNet Classification with Deep Convolutional Neural Networks. In F. Pereira, C. J. C. Burges, L. Bottou, & K. Q. Weinberger (Eds.), *Advances in neural information processing systems* (pp. 1097-1105). Red Hook, NY: Curran Associates, Inc.

Kufareva, I., & Abagyan, R. (2012). Methods of protein structure comparison. *Methods in molecular biology (Clifton, N.J.), 857*, 231-257. doi:10.1007/978-1-61779-588-6_10

Kuncheva, L. I. (2002). A Theoretical Study on Six Classifier Fusion Strategies. *IEEE Trans. Pattern Anal. Mach. Intell., 24*(2), 281-286. doi:10.1109/34.982906

Maeda, k. (2010). From the subspace methods to the mutual subspace method. In *Computer Vision,* (Vol. 285, pp. 135-156). Berlin and Heidelberg: Springer.

Mills, C. L., Beuning, P. J., & Ondrechen, M. J. (2015). Biochemical functional predictions for protein structures of unknown or uncertain function. *Computational and structural biotechnology journal, 13*, 182-191. doi:10.1016/j.csbj.2015.02.003

Mirceva, G., Cingovska, I., Dimov, Z., & Davcev, D. (2012). Efficient approaches for retrieving protein tertiary structures. *IEEE/ACM transactions on computational biology and bioinformatics, 9*(4), 1166-1179. doi:doi:10.1109/TCBB.2011.138

Murzin, A. G., Brenner, S. E., Hubbard, T., & Chothia, C. (1995). SCOP: A structural classification of proteins database for the investigation of sequences and structures. *Journal of Molecular Biology, 247*(4), 536-540. doi:doi:10.1016/S0022-2836(05)80134-2

Nanni, L., & Lumini, A. (2006). An ensemble of K-Local Hyperplane for predicting Protein-Protein interactions. *BioInformatics, 22*(10), 1207-1210.

Nanni, L., Lumini, A., & Brahnam, S. (2014). An empirical study of different approaches for protein classification. *The Scientific World Journal, Article ID 236717*, 1-17. doi:doi:10.1155/2014/236717

Novotny, M., Madsen, D., & Kleywegt, G. J. (2004). Evaluation of protein fold comparison servers. *Proteins: Structure, Function, and Bioinformatics, 54*(2), 260-270. doi:10.1002/prot.10553

O'Donoghue, S. I., Goodsell, D. S., Frangakis, A. S., Jossinet, F., Laskowski, R. A., Nilges, M., . . . Olson, A. J. (2010). Visualization of macromolecular structures. *Nature Methods, 7*(3 Suppl), S42-S55. doi:doi:10.1038/nmeth.1427

Russell, R. B., Russell, R. B., Saqi, M. A., Sayle, R. A., Bates, P. A., & Sternberg, M. J. (1997). Recognition of analogous and homologous protein folds: analysis of sequence and structure conservation. *Journal of Molecular Biology, 269*(3), 423-439. doi:10.1006/jmbi.1997.1019

Sael, L., Li, B., La, D., Fang, Y., Ramani, K., Rustamov, R., & Kihara, D. (2008). Fast protein tertiary structure retrieval based on global surface shape similarity. *Proteins, 72*, 1259-1273. doi:doi:10.1002/prot.22030

Schenkel, A., Holm, L., Rosenström, P., & Kääriäinen, S. (2008). Searching protein structure databases with DaliLite v.3. *BioInformatics, 24*(23), 2780-2781. doi:10.1093/bioinformatics/btn507

Shindyalov, I. N., & Bourne, P. E. (1998). Protein structure alignment by incremental combinatorial extension (CE) of the optimal path. *Protein Engineering, 11*(9), 739-747. doi:doi:10.1093/protein/11.9.739

Simonyan, K., & Zisserman, A. (2014). *Very deep convolutional networks for large-scale image recognition*. Retrieved from arXiv:1409.1556v6

Su, H., Maji, S., Kalogerakis, E., & Learned-Miller, E. (2015). *Multi-view convolutional neural networks for 3d shape recognition*. Paper presented at the 2015 IEEE International Conference on Computer Vision (ICCV), Boston, MA.

Suryanto, C. H., Saigo, H., & Fukui, K. (2015). Structural class classification of 3d protein structure based on multi-view 2d images. *IEEE/ACM transactions on computational biology and bioinformatics, 15*(1), 286-299. doi:doi:10.1109/TCBB.2016.2603987

Szegedy, C., Liu, W., Jia, Y., Sermanet, P., Reed, S., Anguelov, D., . . . Rabinovich, A. (2015). *Going deeper with convolutions*. Paper presented at the IEEE Computer Society Conference on Computer Vision and Pattern Recognition.

Wang, W., & Shen, J. (2018). Deep visual attention prediction. *IEEE Transactions on Image Processing, 27*(5), 2368-2378. doi:doi:10.1109/TIP.2017.2787612

Wu, Z., Song, S., Khosla, A., Yu, F., Zhang, L., Tang, X., & Xiao, J. (2015). *3D ShapeNets: A deep representation for volumetric shapes*. Paper presented at the 2015 IEEE Conference on Computer Vision and Pattern Recognition (CVPR), Boston, MA.

Ye, Y., & Godzik, A. (2003). Flexible structure alignment by chaining aligned fragment pairs allowing twists. *BioInformatics, 19*(Suppl 2), ii246-ii255. doi:doi:10.1093/bioinformatics/btg1086

Yosinski, J., Clune, J., Bengio, Y., & Lipson, H. (2014). *How transferable are features in deep neural networks?* Retrieved from arXiv:1411.1792:

Zeng, Y. H., Guo, Y. Z., Xiao, R. Q., Yang, L., Yu, L. Z., & Li, M. L. (2009). Using the augmented Chou's pseudo amino acid composition for predicting protein submitochondria locations based on auto covariance approach. *Journal of Theoretical Biology, 259*(2), 366-372. doi:doi:10.1016/j.jtbi.2009.03.028

Zhang, C. T., & Chou, K. C. (1994). Predicting protein folding types by distance functions that make allowances for amino acid interactions. *J Biol Chem, 269*(35), 22014-22020.

Zhang, Y., & Skolnick, J. (2005). TM-align: a protein structure alignment algorithm based on the TM-score. *Nucleic Acids Research, 33*(7), 2302-2309. doi:doi:10.1093/nar/gki524

Zhou, G. P. (1998). An intriguing controversy over protein structural class prediction. *Journal of Protein Chemistry, 17*, 729-738.

Zhou, X., Chou, J., & Wong, S. T. C. (2006). Protein structure similarity from principle component correlation analysis. *BMC Bioinformatics, 7*(40). doi:doi:10.1186/1471-2105-7-40